\documentclass[runningheads]{llncs}

\usepackage[colorinlistoftodos]{todonotes}

\usepackage{graphicx}
\usepackage{graphicx,amsfonts,amsmath}
\usepackage{cases}
\usepackage{caption}
\usepackage{subcaption}
\usepackage{comment}
\usepackage{algorithm2e}

\newtheorem{theo}{Theorem}[section]

\newtheorem{defi}{Definition}
\newtheorem{rem}{Remark}

\let\oldexists\exists
\renewcommand{\exists}{\, \oldexists \, }
\let\oldforall\forall
\renewcommand{\forall}{\, \oldforall\, }

\usepackage{graphicx}
\begin{document}
\title{Convergence and scaling of Boolean-weight optimization for hardware reservoirs}
%
%
\author{Louis Andreoli \inst{3}, Stéphane Chrétien \inst{1,2,3}\orcidID{0000-0002-4544-1315} \and Xavier Porte \inst{3} \and Daniel Brunner \inst{3}}
\authorrunning{S. Chrétien et al.}
%
\institute{Université Lyon 2, Laboratoire ERIC, 5 av. Pierre Mendès-France, 69676 Bron, France
\email{stephane.Chretien@univ-lyon2.fr}\\
\url{https://sites.google.com/site/stephanegchretien/} \and
The Alan Turing Institute, British Library, 96 Euston Rd, London NW1 2DB, United Kingdom \\
\and Institut FEMTO-ST, Département AS2M
24, rue Alain Savary
25000 Besançon, France
}
\maketitle              
\begin{abstract}
Hardware implementation of neural network are an essential step to implement next generation efficient and powerful artificial intelligence solutions.
 Besides the realization of a parallel, efficient and scalable hardware architecture, the optimization of the system's extremely large parameter space with sampling-efficient approaches is essential.
 Here, we analytically derive the scaling laws for highly efficient Coordinate Descent applied to optimizing the readout layer of a random recurrently connection neural network, a reservoir.
 We demonstrate that the convergence is exponential and scales linear with the network's number of neurons.
 Our results perfectly reproduce the convergence and scaling of a large-scale photonic reservoir implemented in a proof-of-concept experiment.
 Our work therefore provides a solid foundation for such optimization in hardware networks, and identifies future directions that are promising for optimizing convergence speed during learning leveraging measures of a neural network's amplitude statistics and the weight update rule.
\end{abstract}

\section{Introduction}

With the recent breakthroughs in neural network (NN) algorithms \cite{LeCun2015} came an explosion of research into alternative hardware systems with the aim to realize a new generation of NN processors.
 The general motivation within this field is to create NN processors more adhering to a NN's architecture, which typically is associated with dropping the von Neumann computing concept and to avoid serial rooting where possible \cite{Markovic2020}.
 Efforts investigating such hardware have accelerated on such a scale \cite{Reuther2020} that the term of Cambrian explosion for NN hardware has been coined.
 
As usual, electronics leads the way as it has access to a highly mature integrated technological platform providing advanced design and fabrication technology \cite{Sebastian2018}, \cite{Tsai2018}, \cite{Neckar2019}, \cite{Reuther2020}.
 However, realizing the staggering amount of a NN's connections in parallel makes photonics a technology of continuing appeal.
 Due to the fundamental difference of the information-carrying particle, electronics are to be preferred for information transformation, photons for information transduction \cite{Lohmann1990}.
 Or more elegantly put: “Conventional wisdom says that electrons compute and photons communicate” \cite{Athale2016}.
 Today, parallel photonic interconnects for NNs based on standard 2D integrated photonics \cite{Shen2016,Tait2017}, 3D printed waveguides \cite{Moughames2020} and even continuous media \cite{Psaltis1990,Khoram2019,Dinc2020}.
 
Of particular relevance for the analog hardware neural network revival is the Reservoir Computing (RC) concept \cite{Jaeger2004}.
 The original motivation for RC was a tremendous simplification of optimizing the network's weights since it avoided the classical pit-falls of optimizing recurrent neural networks.
 However, restricting training to the final readout layer simultaneously facilitated implementing such a system in hardware; ultimately it is significantly more straightforward to build some rather than a specific high-dimensional nonlinear system.

Yet, training hardware NNs is a challenge, even in the simplified context of RC.
 Sampling a NN's state is impractical as the probing and signal acquisition hardware would significantly exceed the scale and complexity of the original NN.
 One way mitigating this challenge is in situ learning \cite{Ernoult2020,Wright2022}, another is to train the RC using local modifications according to Coordinate Descent (CD).
 In \cite{Bueno2018,Andreoli2020} we physically implemented Boolean readout weights based on a digital micro-mirror device, whose weights were optimized one at a time per training epoch based on a reward signal computed solely based on the system's output.
 As a weight-update rule we explored either random or greedy descent strategies.

Coordinate Descent is of great interest due to its applicability and superior empirical performance in machine learning and large-scale data analysis problems, such like LASSO-type approaches to sparsity based models such as in \cite{friedman2007pathwise,friedman2010regularization}.
 Beyond the LASSO and its avatars, there has been increased interest in the latest years in the use of CD for large-scale challenges in machine learning, where it has been shown extremely competitive with other methods.
 Prominent examples are the training of linear support vector machines \cite{hsieh2008dual} and non-negative matrix factorization \cite{hsieh2011fast}.
 However, most relevant for the optimizing hardware NN processors is that CD is especially attractive for problems where computing gradients is difficult or even impossible.

In \cite{Bueno2018,Andreoli2020}, we experimentally determined exponential convergence and found that the convergence time scales linear with the system's size.
 However, these findings were limited to the particular task of chaotic signal prediction. 
 Here, we provide the general mathematical framework and show that exponential convergence and a linear scaling learning effort is a fundamental property of Boolean CD, independent of the task.
 Exponential convergence as well as linear scalability are both advantageous properties, and our analytical confirmation of the experimental results provide the basis for broadly applying CD to hardware NNs.
 Finally, we analytically establish the link between NN nodes' amplitude and weight-update statistics, which opens the door for systematically accelerating optimization-convergence during learning.
	
\section{Boolean evolutionary learning \label{sec:LearningExp}}

Details of the hardware system can be found in \cite{Bueno2018}.
 Here, we focus on the aspects most relevant for the general understanding and the mathematical description.
 The state of hardware neuron $i$ at discrete time $n$ is encoded in electromagnetic field $E_{i}(n)$, which at the reservoir's output is converted into
\begin{equation}
y^{out}(k,n)\propto
\sum_{i}^N W_{i}^{\mathrm{DMD}}(k)\left| E_{i}(n)\right|^{2},
\label{eq:Yout}
\end{equation} 
\noindent via an optical detector that provides output signal $y^{\mathrm{out}}(k,n)$, and $N$ is the number of neurons inside the reservoir.
 Each neuron $i$ is assigned Boolean readout weight $W_{i}^{\mathrm{DMD}}(k)$, and training optimizes the Boolean matrix $\mathbf{W}^{\mathrm{DMD}}$ over the duration of the $k =1,\ldots,K$ learning epochs such that output $y^{\mathrm{out}}(K,n)$ best approximates target signal $\mathcal{T}(n)$.
 According to Eq.~\eqref{eq:Yout} training therefore corresponds to the optimization of a second order polynomial function weighted by Boolean coefficients.
 Learning is stopped at epoch $K$ when the hardware NN's output approximates $\mathcal{T}$ within accuracy limit $\epsilon$.
 Finally, for the rest of the paper we refer to the global operation of iteratively tweaking the initial Boolean weights towards the final optimized configuration as the \textit{Boolean minimizer}.
 
\subsection{The greedy learning algorithm} 
\label{sec:greedy}
The Boolean minimizer that was implemented in our experiment can be divided into three functional sections:

\bigskip
\begin{description}
    \item[I. Mutation]
\end{description}
\begin{eqnarray}
\mathbf{W}^{\mathrm{select}}(k) = \mathrm{rand}(N)\cdot \mathbf{W}^{\mathrm{bias}}(k) , \label{eq:Wselect} \\
l(k) = \mathrm{max}(\mathbf{W}^{\mathrm{select}}(k)) , \label{eq:lk} \\
W^{\mathrm{DMD}}_{l(k)}(k+1) = \neg (W^{\mathrm{DMD}}_{l(k)}(k)) , \label{eq:WDMDmodifExploration} \\
\mathbf{W}^{\mathrm{bias}}(k+1) = 1/N + \mathbf{W}^{\mathrm{bias}}(k),  W^{\mathrm{bias}}_{l(k)}=0. \label{eq:Wbias}
\end{eqnarray}

\noindent We create a vector with $N$ independent and identically distributed random elements between 0 and 1 ($\mathrm{rand}(N)$), and $\mathbf{W}^{\mathrm{bias}}$ offers the possibility to modifying the otherwise stochastic selector $\mathbf{W}^{\mathrm{select}}(k) \in \mathbb{R}^N$, Eq.~\eqref{eq:Wselect}.
The largest entry's position in $\mathbf{W}^{\mathrm{select}}(k)$ is $l(k)$, Eq.~\eqref{eq:lk}, which determines the Boolean readout weight $W^{\mathrm{DMD}}_{l(k)}(k)$ to be mutated via a logical inversion (operator $\neg(\cdot)$), see Eq.~\eqref{eq:WDMDmodifExploration}.

A fully stochastic Markovian descent is obtained when replacing $\mathbf{W}^{\mathrm{bias}}$ with a unity matrix and excluding Eq.~\eqref{eq:Wbias}.
However, we also investigate exploration which avoids mutating a particular connection in near succession.
There, $\mathbf{W}^{\mathrm{bias}}$ is randomly initialized at $k=1$.
At each epoch $k$, Eq.~\eqref{eq:Wbias} increases the bias of all connections by $1/N$, while the corresponding bias for the currently modified connection weight is set to zero.
The probability of again probing a particular weight reaches unity after $N$ learning epochs have passed, and we therefore refer to this biased descent as \emph{greedy} learning.

\bigskip
\begin{description}
    \item[2. Error and reward signals]
\end{description}
\begin{eqnarray}
\Phi^{(k)} = \frac{1}{T} \sum_{n=1}^{T}\left(\mathcal{T}(n+1)-\tilde{y}^{\textrm{out}}(k,n+1)\right)^{2},\label{eq:epsilon}  \\
r(k) = \begin{cases}
1 \quad \textrm{if}\ \Delta\Phi^{(k)} < 0\\
0 \quad \textrm{if} \ \Delta\Phi^{(k)} \ge 0
\end{cases}\label{eq:r(k)}. \label{eq:kMin}
\end{eqnarray}

\noindent Mean square error $\Phi^{(k)}$ is obtained from a sequence of $T$ data points according to Eq.~\eqref{eq:epsilon}, and comparison to the previous error assigns a reward $r(k)=1$ only if a modification $\Delta\Phi^{(k)}=\Phi^{(k)} - \Phi^{(k-1)}$ was beneficial, Eq.~\eqref{eq:r(k)}.

\bigskip
\begin{description}   
    \item[3. Descent action]
\end{description}
\bigskip
\begin{equation}\label{eq:WDMDmodifReward}
W^{\textrm{DMD}}_{l(k),k} = r(k)W^{\textrm{DMD}}_{l(k),k}+(1-r(k))W^{\textrm{DMD}}_{l(k),k-1} \ .
\end{equation}

\smallskip
\noindent Based on reward $r(k)$, the DMD's current configuration either accepts or rejects the previous modification, Eq.~\eqref{eq:WDMDmodifReward}.
 Simultaneously modifying groups of DMD mirrors is straight forward in principle, however, we found that convergence in our system and task is slower in that case.

\subsection{Coordinate Descent in hardware \label{sec:ExpResults}}

We have studied the performance and convergence of CD in our photonic RC \cite{Bueno2018,Andreoli2020}.
 We randomly selected an initial Boolean weight configuration $\mathbf{W}^{\mathrm{DMD}}(1)$, which was used as starting point for all descents, and we collected the convergence data of 20 (14) minimizers for greedy (random) descent.
 
\begin{figure}[t]
 \centering
 \includegraphics[width=1\columnwidth]{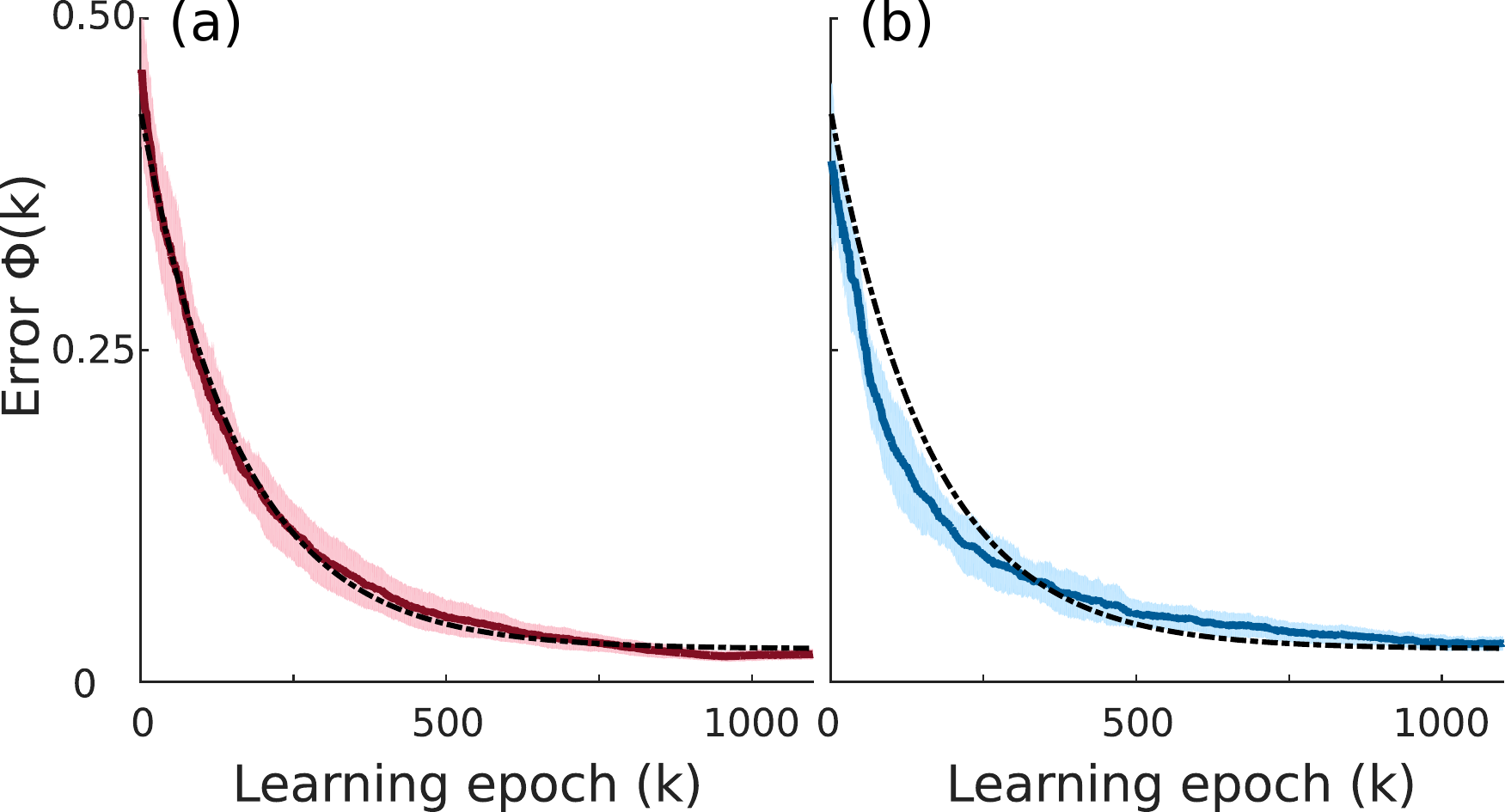}
 \caption{Convergence of Boolean minimizer.
  The mean and standard deviation of Greedy or Markovian exploration are shown in the panel~(a) and (b), respectively.
  The average data is fit with an exponential decay, dash-dotted line, and agreement with greedy exploration is excellent.
  Markovian exploration appears to require additional terms.
  }\label{fig:20learningCurves}
  \vspace{0.05cm}
  \hrule
\end{figure}

Figure~\ref{fig:20learningCurves} (a) and (b) show the convergence obtained for a reservoir of $N=961$ photonic neurons and for greedy and random descent, respectively.
 The dark solid lines are the average error $\bar{\Phi}(k)$, while the colored areas indicate the standard deviation for every epoch $k$.
 A fit of the averaged minimizers to an exponential decay, dashed line in Fig.~\ref{fig:20learningCurves}, agrees remarkably well with the experimentally obtained convergence, in particular for greedy CD.
 We found that convergence of the greedy and random minimizers happens at approximately the same rate, however that the error for greedy optimization $\Phi^{(K)} = (14.2 \pm 2.9)\cdot10^{-3}$ is above the $\Phi^{(K)}=(13.4 \pm 1.9)\cdot10^{-3}$ for random optimization.
 The biggest impact of the greedy descent was found in the time it took both algorithms to converge: for greedy descent best performance was obtained for $K=973.6\pm63.7$, while it took $K=1856.5\pm175.1$ epochs for the random descent to reach its local minimum.

\begin{figure}[t]
 \centering
 \includegraphics[width=0.9\columnwidth]{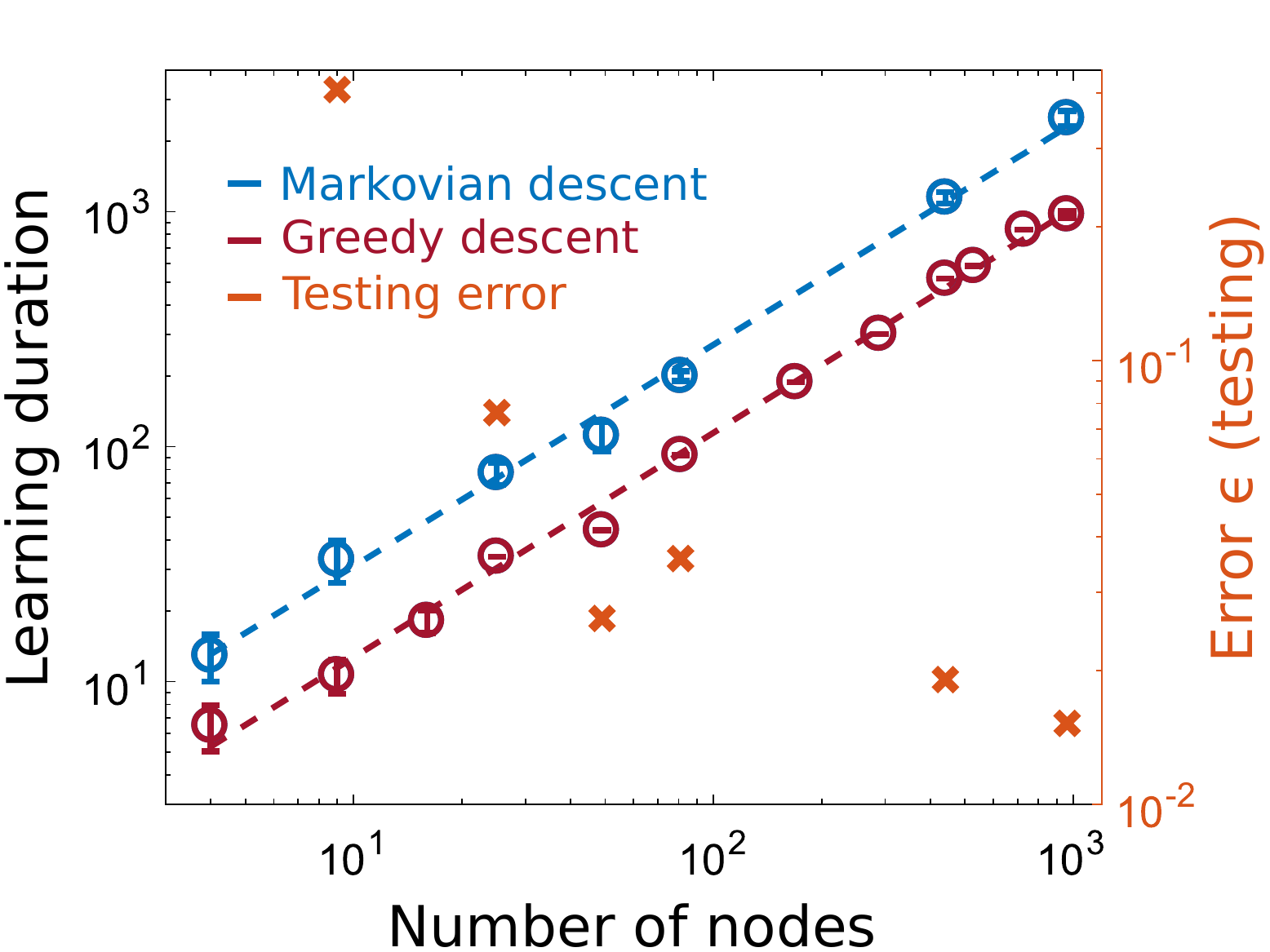}
 \caption{Scaling of convergence speed with system size.
  Greedy and Markovian exploration both result in close to linear scaling with an exponent of $N^{0.96}$ and $N^{0.94}$, respectively.
  Greedy exploration causes approximately twice faster convergence.
  Orange data are testing errors, which continue to decrease.
  }\label{fig:Scaling}
  \smallskip
\hrule
\end{figure}

We then studied $K$ and its dependency on the RC's number of nodes $N$.
 Results for both minimizers are shown in Fig.~\ref{fig:Scaling}.
 The dependency reflects the slower convergence for the random descent, but most importantly we find that the time to reaching a local minima scales linear with the system's size.
 We have confirmed this finding across three orders of magnitude, which in combination with the statistics obtained from the numerous minimizers makes this a robust experimental observation.
 Such linearly scaling is of great importance to future generations of hardware implemented RC, and a general mathematical confirmation is required.

\section{Random Coordinate Descent methods}

The Boolean evolutionary learning algorithm of the previous section belongs to the class of random Coordinate Descent algorithms, which is currently undergoing a period of tremendous research activity in the mathematics community.
 In the field of continuous optimisation in particular, CD algorithms have quite a long  history and recently gained interest in the applied machine learning community for handing big data analytics problems.
 They are iterative methods in which each iterate is obtained by fixing most components of the variable vector $x$ at their current iteration's values, and approximately minimizing the objective, i.e. error $\Phi$.

 Concerning its theoretical convergence properties, they have been extensively studied in the 80s and 90s \cite{ortega2000iterative}, \cite{bertsekas1989parallel}, \cite{luo1992convergence} and \cite{luo1993error}.
 More recently, Nesterov \cite{nesterov2012efficiency} provided the first global non-asymptotic convergence rates for the method when applied to convex and smooth problems; see also Beck and Tetruashvili \cite{beck2013convergence} for block coordinate gradient descent with the cyclic schedule.
 Tighter rate estimates for the cyclic scheme have been presented in \cite{sun2015improved}. 

In comparison, the case of optimization over combinatorial domains has been little studied.
 An important body of work around random methods, called evolutionary algorithms was done by Doerr \cite{doerr2020probabilistic}, Wegener \cite{wegener2005complexity} and many other authors.
 We were not, however, able to use their results for the analysis of our Algorithm introduced in Section \ref{sec:LearningExp}.
 Closer to our work is the paper \cite{yuan2017hybrid} which addresses the case of composite functions is penalised regression. 

\section{Analysis of the algorithm}
The introduced binary optimization problem can be rewritten in the matrix form 
\begin{align*}
    \min_{W^{\mathrm{DMD}} \in \{0,1 \}^N} \ \Vert \mathcal T- \mathcal E W^{\mathrm{DMD}}\Vert_2^2
\end{align*}
with
\begin{align}
    \mathcal E & = 
    \begin{bmatrix}
         \vert E_1(2) \vert^2 & \cdots & \vert E_N(2)\vert\\
         \vdots &                      & \vdots \\
         \vert E_1(T+1) \vert^2 & \cdots & \vert  E_N(T+1) \vert^2\\
    \end{bmatrix}.
    \label{eq:Ematrix}
\end{align}
Using the change of variable 
\begin{align*}
    x_i & = 2W^{\mathrm{DMD}}_{i}-1       
\end{align*}
for $i=1,\ldots,N$, 
the problem can be rewritten as the one of solving
	\begin{align}
	\min_{x \in \{-1,1\}^N  } \ 
	\Phi(x) :=\Vert \mathcal T-\frac12 \mathcal E \mathbf{1} -\frac12 \mathcal Ex\Vert_2^2, 
	\end{align}
where $\mathbf{1}$ denotes the vector of all ones.
Notice that, being quadratic, $\Phi$ is automatically $\lambda$-smooth with $\lambda:=\lambda_{\max} (\mathcal E^t\mathcal E)$ as the largest eigenvalue of matrix $\mathcal E^t \mathcal E$.
This means that the gradient of $\Phi$ is Lipschitz continuous, i.e. $\vert \nabla \Phi(x)-\nabla \Phi(x')\vert \le \lambda \Vert x-x'\Vert_2$ for any $x$, $x'$ in $\mathbb R^N$, a stronger requirement than mere continuity.

According to Sec. \ref{sec:greedy} the update rule for $x^{(k)}$ at optimization epoch $k$ depends on the realisation $i$ of a random index variable taking values in $\{1,\ldots,N\}$ distributed according to probability mass $\pi$. 
	
	\begin{defi}
		A binary vector $\bar x$ is said  to be a local coordinatewise minimiser if 
		\begin{align*}
		\Phi(x') & \ge \Phi(x)
		\end{align*} 
		for all binary vectors $x'$ that differ from $x$ by only one coordinate. 
	\end{defi}
	Notice further that the constant $C$ can be arbitrary because for any $C$, $\Phi$ has the same coordinate-wise local minimisers.

\subsection{The main algorithm}

The main algorithm we will study works as follows:

\begin{algorithm}
		\SetAlgoLined
		\KwResult{The final iterate $x^{(K)}$}
		Choose $x^{(0)}$ at random\;
		\While{$x^{(k)}$ is not a local coordinate minimiser}{
			Choose a coordinate $i$ at random with probability $\pi_i^{(k+1)}$\;
			Change $x_i^{(k)}$ to its complementary value $\tilde x_i = -x_i^{(k)}$\;
			For all $i'\neq i$, set $\tilde x_{i'} = x_{i'}$\;
			\eIf{$\Vert y-A\tilde x\Vert_2^2 < \Vert y-A x\Vert_2^2$}{
				$x^{(k+1)} = \tilde x$\;
			}
			{$x^{(k+1)}=x^{(k)} $}
		}
		\caption{Greedy learning with binary weights}
		\label{alg}
	\end{algorithm}
	
	\subsection{Main result}

	\begin{rem}
	\label{autostrcvx}
	Notice that we can always assume that the function $\Phi$ is equal to 
	\begin{align}
	\Phi(x)= \Vert a-Ax\Vert_2^2+C
	\end{align} 
	with $a=\mathcal T-\frac12 \mathcal E\mathbf{1}$, $A=\frac12 \mathcal E$,
	for any positive constant $C$.
	Using this trick, we can also assume that $\Phi$ is strongly convex, i.e. converges faster than quadratically: for twice differentiable functions, strong convexity is equivalent to the Hessian's smallest eigenvalue being bounded from below by a positive (nonzero) constant.
	Since we restrict our analysis to the case of binary vectors, we can add a constant term $\frac{\eta}2 \Vert x\Vert_2^2=\frac{\eta}2 N$, ergo satisfying this condition for strong convexity.
	\end{rem}

The following definition is important for our analysis.
 Our update rule for computing $x^{(k+1)}$ from $x^{(k)}$ depends on value $i$ of a random index variable $I$ taking values on $\{1,\ldots,N\}$ with probability mass function $\pi^{(k+1)}$.
 Define $\Pi$ as the rounding operator 
\begin{align}
	\Pi(a) & = \text{argmin}_{x \in \{-1,1\}^N} \Vert a-x\Vert_2^2 .
\end{align}
Define the quantity $\kappa$ as 
 \begin{align}
 \label{kappa}
 & \kappa = 1- &  \\
 & \max_{\stackrel{x \in \{-1,1\}^N}{x' \neq \Pi(x), \pi \in \Delta_N}}  \frac{\Vert D(\sqrt{\pi}) ( \Pi(x-\frac{1}{\lambda} \nabla \Phi(x)) -x+\frac{1}{\lambda} \nabla \Phi(x)) \Vert_2}{\Vert D(\sqrt{\pi}) ( x' -x+\frac{1}{\lambda} \nabla \Phi(x))\Vert_2}.&
 \nonumber 
 \end{align}
 Notice that this quantity is larger than zero for generic problems, due to the constraint $x' \neq \Pi(x)$ in the optimisation. 

	Our main result is the following.	
	\begin{theo}
	\label{main}
		Let $x^{(k)}$, $k=1,\ldots K$ be defined by Algorithm \ref{alg}. 
		Let us assume that $x^{(k)}$, $k=1,\ldots K$ converges to $\bar x\in \{-1,1\}^N$.
		Assume that $x^{(k)} \neq \bar x$ for all $k=1,\ldots,K$ and let  $\lambda$ denote the largest eigenvalue of $\mathcal E^t\mathcal E$. Then, we have 
		\begin{align}
	        \mathbb E \left[\Phi(x^{(k+1)}) - \Phi(\bar x)\right] & \le 
	        \rho \ \mathbb E \left[\left(\Phi(x^{(k)})-\Phi(\bar x)\right)\right] \nonumber
	\end{align}
	with 
	\begin{align*}
	    \label{rho}
	    \rho & = \kappa \left(1-\Vert \pi^{(k+1)}\Vert_{\infty}\frac{\lambda(1-\kappa)}{\eta} \  \Big(\frac{\eta}{2N} \  -1\Big)\right).
	\end{align*}
	\end{theo}

Notice that, by construction, we know that 
        \begin{align*}
            \Phi(x^{(k+1)}) & \le \Phi(x^{(k)})
        \end{align*}
for all $k=1,\ldots,+\infty$ since we only update $x^{(k+1)}$ if the error was reduced.
 This ensures that the successive iterates improve the loss value at every iteration, or keep it constant.
 The algorithm can of course stop in finite time.
 Notice moreover that we can analyse the number $K$ of iterations it needs to achieve an accuracy of order $\epsilon$, $\rho^K \sim \epsilon$, i.e. $$K \sim \log (1/\epsilon)/\log(1/\rho).$$
 In particular, if we take $\eta$ sufficiently large, we have
	\begin{align}
	    & \Vert \pi^{(k+1)}\Vert_{\infty}\frac{\lambda(1-\kappa)}{\eta} \  \Big(\frac{\eta}{2N}  -1\Big)\\
	    & \hspace{1cm}\sim \Vert \pi^{(k+1)}\Vert_{\infty}\frac{\lambda(1-\kappa)}{2N}.    \nonumber 
	\end{align}	
Moreover, if descent coordinates are selected according to probability distribution with maximum entry $\Vert \pi^{(k+1)}\Vert_{\infty} \sim \tau /N^\alpha$ for some non-negative $\alpha$.
 The uniform case for random CD corresponds to $\alpha=1$, then the number $K$ of iterations in order to achieve convergence up to a tolerance $\epsilon>0$ is of the order 
	\begin{align}
	    & \log(\epsilon^{-1})  -\log (1-\lambda/( N^{1+\alpha}))) \nonumber \\
	    & \hspace{.5cm} \sim \log(\epsilon^{-1})  \ \lambda/(N^{1+\alpha}).
	\end{align}
Now, if $\lambda \sim N^\beta$, which is the case with $\beta=1$ for our random update policy, i.e. if $\mathcal E$ is a random matrix with i.i.d. centered subGaussian entries and $N\ge T$ then we get that 
	\begin{align}\label{eq:scaling}
	    K & \sim \log(\epsilon^{-1}) N^{1+\alpha-\beta}.
	\end{align}

\section{\label{sec:Discussion} Discussion}

Our Theorem \ref{main} and Eq.~\ref{eq:scaling} show that the method converges at an exponential speed on average to a point $\bar x$.
 This corroborates the linear $\log-\log$ dependency on the dimensionality illustrated in Fig.~\ref{fig:Scaling}.
 We observe that Eq.~\eqref{eq:scaling} depends on, both, the update rule $\pi^{(k)}$ through $\alpha$ as well as the amplitude distribution of network nodes through $\beta$.
 The case of uniform CD, i.e. each component is drawn independently uniformly at random and $\alpha=1$.

Concerning the value of the parameter $\beta$, several previous results on random matrix theory give us an interesting answer.
It is well known that random matrices with i.i.d. subGaussian entries satisfy $\lambda \sim (\sqrt{N}+\sqrt{T})^{2}$; see for instance \cite{vershynin2018high}.
This is true in particular for the uniform, Gaussian, and many other distributions. When $T$ is of the same order as $N$, we get that $\beta \sim 1$.
The correct order of $\alpha$ seems much trickier to derive using a purely analytical arguments and we resort to an empirical approach. 

We can infer $\alpha - \beta$ from the curves.
 To obtain a better overview of both contributions, in the future we can independently modify update rule $\pi^{(k)}$ to explore its impact on $\alpha$, while keeping the network's amplitude distribution fixed, i.e. under identical conditions for $\beta$.
 Vice-versa, in the experiment we can readily modify the network's amplitude distribution and hence $\beta$, while keeping $\alpha$ constant.
 Following these strategies, we should be able to further experimentally investigate the relationship between the system, its convergence and $\alpha$.

\section{\label{sec:Conclusion} Conclusion}


Previously, we implemented random coordinate descent in a Boolean configuration space in a large scale photonic NN, with the motivation for creating a proof-of-concept hardware concept that is scalable and not limited by the von Neumann bottleneck.
 Here, we analytically show that cost-function minimization converges exponentially for such a optimization routine, and most importantly that the time of convergence scales linear with the system's size.
 This finding we experimentally confirm for different descent strategies based on a photonic reservoir comprising 961 neurons and Boolean readout weights implemented in an array of micro-mirrors.


Our analytical results and the experimental confirmation are important for future hardware implemented NNs.
 Boolean weights are relevant in photonic \cite{Andreoli2020} as well as in electronic \cite{Hirtzlin2020} NN implementations.
 The reported linear scaling ensures scalability of the learning concept, which is an important advantage considering the expected inferior performance in terms of inference accuracy when compared to error back propagation.
 Finally, binary or trinary weight current experience significant interest due to their energy and hardware efficient implementation, and due to the possibility of implementing learning without forgetting when their are the rounded and clipped version of an internal, continuous variable that is optimized with continuous gradients \cite{Laborieux2021}.

\section{Acknowledgment}

The authors acknowledge the support of the Region Bourgogne Franche-Comté.
 This work is supported by the EUR EIPHI program (Contract No. ANR-17-EURE- 0002), by the Volkswagen Foundation (NeuroQNet I), by the French Investissements d’Avenir program, project ISITE-BFC (contract ANR-15-IDEX-03). 
 Author X. Porte receives funding from the Marie Skłodowska-Curie grant agreement No. 713694 (MULTIPLY).
	
\appendix[Proof of Theorem \ref{main}]
For all $l=1,\ldots K$, the notation $\mathbb E_{l}$ will denote the expectation conditional on $x^{(1)}$,\ldots,$x^{(l)}$.
 
        By construction, we know that 
        \begin{align*}
            \Phi(x^{(l+1)}) & \le \Phi(x^{(l)})
        \end{align*}
        for all $l=1,\ldots,+\infty$. This inequality does however not prove that 
        the sequence converges to a local coordinate minimiser nor yields a rate of convergence.
        In order to obtain a more precise result, we need to leverage smoothness and strong convexity of $\Phi$ which are both natural in problem under study.
		Let $z$ denote a vector in $\{-1,1\}^N$ such that $z_{I^{(l)^c}}=x_{I^{(l)^c}}^{(l)}$, where $I^{(l)^c}=\{1,\ldots,N\}\setminus \{I^{(l)}\}$.
        Using the $\lambda$-smoothness property 
		\begin{align}
		\Phi(z) & \le \Phi(x^{(l)}) + \langle \nabla \Phi(x^{(l)}),z-x^{(l)}\rangle + \frac{\lambda}{2} \ \Vert z-x^{(l)}\Vert_2^2 \nonumber 
		\end{align}
		and the fact that $\Phi(x^{(l+1)}) \le \Phi(z)$, 
        we get  
		\begin{align}
		  \mathbb E_{l} \ \left[\Phi(x^{(l+1)})\right] & \le   \mathbb E_{l} \Bigg[\Phi(x^{(l)}) +\langle \nabla \Phi(x^{(l)}),z-x^{(l)}\rangle + \frac{\lambda}{2} \Vert z-x^{(l)}\Vert_2^2\Bigg].\nonumber
		\end{align}
        Moreover, since $z_{I^{(l)^c}}=x_{I^{(l)^c}}^{(l)}$, we get 
		\begin{align}
		  \mathbb E_{l} \ \left[\Phi(x^{(l+1)})\right] & \le \Phi(x^{(l)}) +  \mathbb E_{l} \Bigg[\langle \nabla \Phi(x^{(l)})_{I^{(l)}},z_{I^{(l)}}-x^{(l)}_{I^{(l)}}\rangle \nonumber\\
		& \quad \quad + \frac{\lambda}{2} \Vert z_{I^{(l)}}-x^{(l)}_{I^{(l)}}\Vert_2^2\Bigg]. 		\label{toto} 
		\end{align}
		Moreover, computing the expectation using $\pi^{(l+1)}$ easily gives 
		\begin{align}
			&   \mathbb E_{l} \left[\langle \nabla \Phi(x^{(l)})_{I^{(l)}},z_{I^{(l)}}-x^{(l)}_{I^{(l)}}\rangle + \frac{\lambda}{2} \Vert z_{I^{(l)}}-x^{(l)}_{I^{(l)}}\Vert_2^2\right] \nonumber \\
			& \quad = \langle \nabla \Phi(x^{(l)}), D(\pi^{(l+1)})(z-x^{(l)})\rangle \\
			& \hspace{2cm}+ 
			\frac{\lambda}{2} \ \Vert D(\sqrt{\pi^{(l+1)}}) (z-x^{(l)})\Vert_2^2\nonumber
		\end{align}
		where $z$ is the vector whose components are $z_1,\ldots,z_N$. Set 
		\begin{align}
		    & z= \Pi\Bigg(x^{(l)}-\frac{\mu}{\lambda}\nabla \Phi(x^{(l)}) \Bigg) \nonumber
		\end{align}
		and obtain 
		\begin{align*}
			&   \mathbb E_{l} \left[\langle  \nabla \Phi(x^{(l)})_{I^{(l)}},z_{I^{(l)}}-x^{(l)}_{I^{(l)}}\rangle + \frac{\lambda}{2} \Vert z_{I^{(l)}}-x^{(l)}_{I^{(l)}}\Vert_2^2\right]  \\
			& \hspace{.5cm} =
			\langle D(\pi^{(l+1)})\nabla \Phi(x^{(l)}), 
			\Pi\left(x^{(l)}-\frac{1}{\lambda}\nabla \Phi(x^{(l)}) \right)-x^{(l)}\rangle \nonumber \\
			& \hspace{.5cm}+ 
			\frac{\lambda}{2} \ \Bigg\Vert D(\sqrt{\pi^{(l+1)}}) \Bigg(\Pi\Big(x^{(l)}-\frac{1}{\lambda}\nabla \Phi(x^{(l)}) \Big) -x^{(l)}\Bigg)\Bigg\Vert_2^2,\nonumber
		\end{align*}
		which gives
		\begin{align}
			&   \mathbb E_{l} \Bigg[\langle D(\pi^{(l+1)})\nabla \Phi(x^{(l)})_{I^{(l)}},z_{I^{(l)}}-x^{(l)}_{I^{(l)}}\rangle \label{toti}  + \frac{\lambda}{2} \Vert z_{I^{(l)}}-x^{(l)}_{I^{(l)}}\Vert_2^2\Bigg] \nonumber \\
			& \le \Big\langle D(\sqrt{\pi^{(l+1)}})\nabla \Phi(x^{(l)}), D(\sqrt{\pi^{(l+1)}})
			\Big(\Pi\Big(x^{(l)} -\frac{\mu}{\lambda}\nabla \Phi(x^{(l)}) \Big)-x^{(l)}\Big)\Big\rangle  \nonumber\\
			& \hspace{-.1cm}+ 
			\frac{\lambda}{2} \ \Bigg\Vert D(\sqrt{\pi^{(l+1)}})  \Bigg(\Pi\left(x^{(l)}-\frac{\mu}{\lambda}\nabla \Phi(x^{(l)}) \right) -x^{(l)}\Bigg)\Bigg\Vert_2^2.\nonumber
		\end{align}
		On the other hand, the Al-Kashi identity
		$$\langle g,y\rangle+\frac{t}{2} \Vert y\Vert_2^2=\frac{t}2\Vert y+t^{-1}g\Vert_2^2- \frac{1}{2t} \Vert g\Vert_2^2$$
		applied to $t=\lambda$, gives
		$$ y =D(\sqrt{\pi^{(l+1)}})\Big( \Pi\left(x^{(l)}-\frac{1}{\lambda}\nabla \Phi(x^{(l)}) \right)-x^{(l)}\Big),$$
		and 
		$$ g = D(\sqrt{\pi^{(l+1)}})\Big(\nabla \Phi(x^{(l)})\Big),$$
		in the right hand side term of Eq.~\eqref{toti} gives
		\begin{align*}
		    &   \mathbb E_{l} \Bigg[\langle D(\pi^{(l+1)})\nabla \Phi(x^{(l)})_{I^{(l)}},z_{I^{(l)}}-x^{(l)}_{I^{(l)}}\rangle  + \frac{\lambda}{2} \Vert z_{I^{(l)}}-x^{(l)}_{I^{(l)}}\Vert_2^2\Bigg] \\
			& = \frac{\lambda}{2} \  \Bigg\Vert D(\sqrt{\pi^{(l+1)}}) \Big(\Pi\left(x^{(l)}-\frac{1}{\lambda}\nabla \Phi(x^{(l)}) \right) -x^{(l)}+ \frac{1} {\lambda} \nabla \Phi(x^{(l)})\Big)\Bigg\Vert_2^2 \\
			& \hspace{1cm} -\frac{1} {2 \ \lambda} \Vert D(\sqrt{\pi^{(l+1)}})\nabla \Phi(x^{(l)})\Vert_2^2.
		\end{align*}
	Now using definition Eq.~\eqref{kappa} into this last equation gives 
	\begin{align*}
	    &   \mathbb E_{l} \left[\langle \nabla \Phi(x^{(l)})_{I^{(l)}},z_{I^{(l)}}-x^{(l)}_{I^{(l)}}\rangle + \frac{\lambda}{2} \Vert z_{I^{(l)}}-x^{(l)}_{I^{(l)}}\Vert_2^2\right] \\ 
	    & \hspace{.2cm} \le \frac{  \lambda(1-\kappa)}{2} \  \Big\Vert   D(\sqrt{\pi^{(l+1)}})\Big(\bar x-x^{(l)} \\
	    & \hspace{.5cm}+ \frac{1} {\lambda} \nabla \Phi(x^{(l)})\Big)\Big\Vert_2^2 -\frac{1} {2 \ \lambda} \Vert D(\sqrt{\pi^{(l+1)}})\nabla \Phi(x^{(l)})\Vert_2^2.
	\end{align*}
	Now, expand the first term in the RHS and obtain
	\begin{align*}
	    & \left\Vert D(\sqrt{\pi^{(l+1)}}) \left(\bar x-x^{(l)}+ \frac{1} {\lambda} \nabla \Phi(x^{(l)})\right)\right\Vert_2^2\le \Vert D(\sqrt{\pi^{(l+1)}}) \left( x-x^{(l)}\right)\Vert_2^2 \\
	    & \hspace{.7cm} +\frac{2} {\lambda} \langle \bar x-x^{(l)},D(\pi^{(l+1)})\nabla \Phi(x^{(l)})\rangle + \frac{1} {\lambda^2}\ \Vert D(\sqrt{\pi^{(l+1)}}) \nabla \Phi(x^{(l)}\Vert_2^2.
	\end{align*}
	Using that, by Remark~\ref{autostrcvx}, $\Phi$ is $\eta$-strongly convex, we get that 
	\begin{align*}
	    \langle \bar x-x^{(l)},D(\pi^{(l+1)})\nabla \Phi(x^{(l)})\rangle & \le \Phi(\bar x)-\Phi(x^{(l)}) - \frac{\eta}2 \ \left\Vert D(\pi^{(l+1)}) \Big(\bar x-x^{(l)} \Big)\right\Vert_2^2. 
	\end{align*}
	Therefore, we get 
	\begin{align*}
	    &   \mathbb E_{l} \left[\langle \nabla \Phi(x^{(l)})_{I^{(l)}},z_{I^{(l)}}-x^{(l)}_{I^{(l)}}\rangle + \frac{\lambda}{2} \Vert z_{I^{(l)}}-x^{(l)}_{I^{(l)}}\Vert_2^2\right]  \le  \frac{\lambda(1-\kappa)}{2 } \  \Big( \Vert D(\sqrt{\pi^{(l+1)}}) \left( x^{(l)}-\bar x \right)\Vert_2^2 \\
	    & \hspace{2cm}-  \frac{\eta}2 \ \Vert D(\pi^{(l+1)}) \left( x^{(l)}-\bar x \right)\Vert_2^2\Big) \\
	    & \hspace{2cm} + (1-\kappa) \Big(\Phi(\bar x)-\Phi(x^{(l)})\Big) 
	    + \frac{1-\kappa} {2\lambda}\ \Vert D(\sqrt{\pi^{(l+1)}}) \nabla \Phi(x^{(l)})\Vert_2^2\\
	    & \hspace{2cm}-\frac{1} {2 \ \lambda} \Vert D(\sqrt{\pi^{(l+1)}})\nabla \Phi(x^{(l)})\Vert_2^2. \nonumber 
	 \end{align*}
	 Using that $\kappa \ge 0$, the two last terms in the right hand side of this last equation combine into a
	 nonpositive term, and as a result, we get  
	 \begin{align*}
	        &   \mathbb E_{l} \left[\langle \nabla \Phi(x^{(l)})_{I^{(l)}},z_{I^{(l)}}-x^{(l)}_{I^{(l)}}\rangle + \frac{\lambda}{2} \Vert z_{I^{(l)}}-x^{(l)}_{I^{(l)}}\Vert_2^2\right] \\ 
	        & \hspace{.2cm} \le \Vert \pi^{(l+1)}\Vert_{\infty}\frac{\lambda(1-\kappa)}{2 } \  \Big(1-\frac{\eta}{2N}\ \Big)\Vert x^{(l)}-\bar x\Vert_2^2 \\
            & \hspace{1cm}+ (1-\kappa) \Big(\Phi(\bar x)-\Phi(x^{(l)})\Big)
	 \end{align*}
	 where 
	 Finally, recall that $\eta$-strong convexity implies that 
	\begin{align}
	    \Vert x^{(l)}-\bar x\Vert_2^2  & \le \frac{2}{\eta} \ (\Phi(x^{(l)})-\Phi(\bar x)).
	\end{align}
    Combining this with Eq.~\eqref{toto}, we get 
	\begin{align*}
	      \mathbb E_{l} \ \left[\Phi(x^{(l+1)})\right] - \Phi(x^{(l)}) & \le 
	    (1-\rho) \left(\Phi(\bar x)-\Phi(x^{(l)})\right)
	\end{align*}
	and thus
	\begin{align*}
	      \mathbb E_{l} \ \left[\Phi(x^{(l+1)})\right] - \Phi(\bar x) & \le 
	    \rho \left(\Phi(x^{(l)})-\Phi(\bar x)\right).
	\end{align*}
Taking expectations on both sides, we get the result.

\bibliographystyle{plain}
\bibliography{output}

\end{document}